\documentclass[letterpaper]{article} 
\usepackage{aaai25}  
\usepackage{times}  
\usepackage{helvet}  
\usepackage{courier}  
\usepackage[hyphens]{url}  
\usepackage{graphicx} 
\usepackage{natbib}  
\usepackage{caption} 
\usepackage{algorithm}

\usepackage{newfloat}
\usepackage{listings}
\DeclareCaptionStyle{ruled}{labelfont=normalfont,labelsep=colon,strut=off} 
\floatstyle{ruled}
\newfloat{listing}{tb}{lst}{}
\floatname{listing}{Listing}
\usepackage{paperpkg}
\usepackage{booktabs}       
\usepackage{algpseudocode}  
\usepackage{subcaption}     

\newtheoremstyle{nodecimal} 
  {3pt} 
  {3pt} 
  {\itshape} 
  {} 
  {\bfseries} 
  {} 
  {.5em} 
  {} 

\theoremstyle{nodecimal}
\newtheorem{theorem}{Theorem}
\newtheorem{proposition}{Proposition}

\def\spacingset#1{\renewcommand{\baselinestretch}%
{#1}\small\normalsize}

\title{Optimized Gradient Clipping for Noisy Label Learning}
\author{
    Xichen Ye\textsuperscript{\rm 1}, 
    Yifan Wu\textsuperscript{\rm 1,2}, 
    Weizhong Zhang\textsuperscript{\rm 2,4}, 
    Xiaoqiang Li\textsuperscript{\rm 1}\thanks{Corresponding authors.}, 
    Yifan Chen\textsuperscript{\rm 3}\footnotemark[1],
    Cheng Jin\textsuperscript{\rm 2,5}
}
\affiliations{

    \textsuperscript{\rm 1}Shanghai University\\
    \textsuperscript{\rm 2}Fudan University\\
    \textsuperscript{\rm 3}Hong Kong Baptist University\\
    \textsuperscript{\rm 4}Shanghai Key Laboratory of Intelligent Information Processing\\
    \textsuperscript{\rm 5}Shanghai Collaborative Innovation Center of Intelligent Visual Computing\\
    \{yexichen0930, victorwu, xqli\}@shu.edu.cn,
    \{weizhongzhang, jc\}@fudan.edu.cn, yifanc@hkbu.edu.hk

}

\usepackage{bibentry}

\begin{document}

\maketitle

\begin{abstract}
Previous research has shown that constraining the gradient of loss function w.r.t.\ model-predicted probabilities can enhance the model robustness against noisy labels.
These methods typically specify a fixed optimal threshold for gradient clipping through validation data to obtain the desired robustness against noise.
However, this common practice overlooks the dynamic distribution of gradients from both clean and noisy-labeled samples at different stages of training, significantly limiting the model capability to adapt to the variable nature of gradients throughout the training process.
To address this issue, we propose a simple yet effective approach called Optimized Gradient Clipping (OGC), which dynamically adjusts the clipping threshold based on the ratio of noise gradients to clean gradients after clipping, estimated by modeling the distributions of clean and noisy samples. 
This approach allows us to modify the clipping threshold at each training step, effectively controlling the influence of noise gradients.
Additionally, we provide statistical analysis to certify the noise-tolerance ability of OGC.
Our extensive experiments across various types of label noise, including symmetric, asymmetric, instance-dependent, and real-world noise, demonstrate the effectiveness of our approach.
\end{abstract}

%
\begin{links}
    \link{Code}{https://github.com/Virusdoll/OGC}
\end{links}

\section{Introduction}

The effectiveness of supervised deep learning relies heavily on the availability of large-scale, qualifiedly-annotated data.
Research has shown that an over-parameterized Deep Neural Network (DNN) can easily fit a dataset with randomly assigned labels \cite{DBLP:conf/iclr/ZhangBHRV17:generalization}, underscoring the importance of high-quality annotations.
However, in real-world scenarios, the annotation process will inevitably introduce noisy labels (incorrectly annotated labels), which is liable to compromise the performance of the model.
Therefore, the study of noisy label learning has attracted significant attention.
The goal of noisy label learning is to train a model on a given corrupted dataset, which contains mislabeled samples, while mitigating the adverse effects of noisy labels, and ultimately enabling the trained model to generalize well on a clean evaluation set.

\begin{figure*}
    \centering
    \includegraphics[width=1\textwidth]{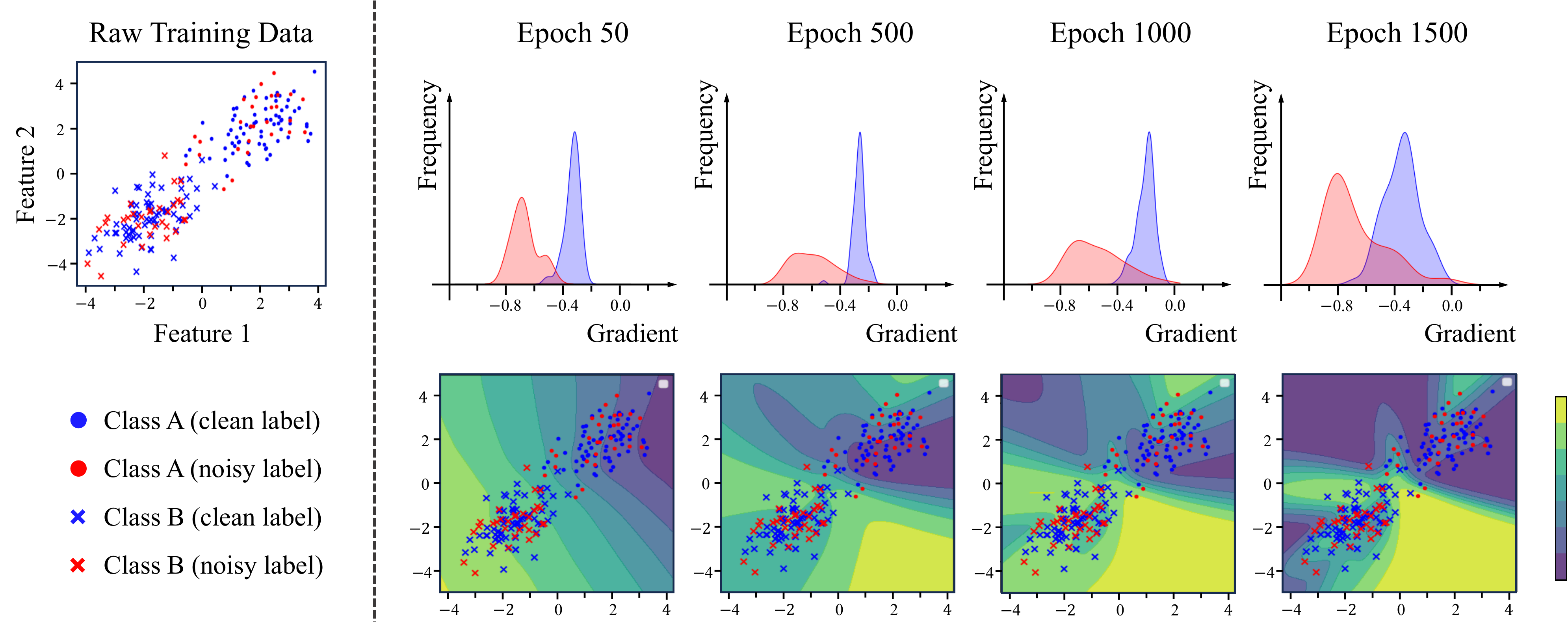}
    \caption{KDE visualizations of gradient distributions for clean and noisy labels, along with decision boundary visualizations for a simple binary classification task that utilizes gradient clipping with a fixed threshold. The leftmost plot in the first row shows the raw training data. The subsequent plots illustrate shifts in gradient distributions (before clipping) at various training epochs (50, 500, 1000, and 1500). The second row displays the corresponding decision boundaries for each epoch.}
    \label{fig: motivation}
\end{figure*}

Cross Entropy (CE), the most widely used loss function for classification, is nevertheless vulnerable to label noise.
In contrast, symmetric loss functions like Mean Absolute Error (MAE) are noise tolerant but less effective for training classification models \cite{DBLP:conf/aaai/GhoshKS17:RobustLoss, DBLP:conf/nips/ZhangS18:GCE}.
This discrepancy has driven the studies into robust loss functions, which primarily focus on how to make full use of both CE and MAE \cite{DBLP:conf/nips/ZhangS18:GCE, DBLP:conf/ijcai/FengSLL0020:Taylor-CE, DBLP:conf/nips/EnglessonA21:GJSLoss, DBLP:conf/iccv/0001MCLY019:SCE, DBLP:conf/icml/MaH00E020:NormalizedLoss, DBLP:conf/nips/YeLD0ST23:NormalizedNegativeLoss} and have achieved notable success.
Besides these, another promising approach, clipping-based techniques, also warrants greater attention.
Among them, \citet{DBLP:conf/iclr/MenonRRK20:PHuberCE} first introduced the PHuberCE method, which refines traditional gradient clipping by applying gradient clipping to model-predicted probabilities (see the definition in Section~\ref{sec:risk-noise}), offering an improved approach to handling noisy labels.
More recently, \citet{DBLP:conf/icml/WeiZXF00L23:LogitClip} proposed LogitClip, which involves clipping the logits to indirectly constrain the upper and lower bounds of the gradients.

Specifically, the two aforementioned clipping-based methods rely on the validation set to determine a fixed optimal clipping threshold, overlooking the dynamic nature of gradients during the training process. 
However, as demonstrated by the simple binary classification experiment in Figure~\ref{fig: motivation}, we observe distinct shifts in the gradient distribution throughout training, which significantly affect the suitability of a fixed threshold.
Considering the fact that employing a higher threshold can enhance the model’s fitting ability, yet fails to bound the decreasing noisy gradients; conversely, employing a strict threshold ensures the noisy gradients remain bounded but compromises the model’s fitting ability, we expect a dynamically adjustable threshold for gradient clipping that continuously adapts in response to the shifts in the gradient distribution. 

Motivated by this, in this paper, we propose an optimize-based strategy, Optimized Gradient Clipping (OGC), which ensures that the threshold adjusts itself as the distribution of gradients evolves during training, maintaining an optimal balance between fitting ability and noise suppression.
Specifically, OGC 1) utilizes a 2-component Gaussian Mixture Model (2-GMM) to model the distribution of cross-entropy losses for clean and noisy samples, 2) and estimates the ratio of the noisy gradients to the clean gradients after clipping.
This allows us to determine a clipping threshold at each training step that restricts the gradient ratio to a predefined limit, effectively bounding the influence of noisy gradients.
Theoretically, we demonstrate that the dynamic threshold obtained by OGC can effectively enhance the noise-tolerant ability of CE under either symmetric, asymmetric, or instance-dependent noise, which is originally a non-robust loss function.
Furthermore, we conduct extensive experiments on various types of label noise, including, symmetric, asymmetric, instance-dependent, and real-world noise, to verify the effectiveness of our proposed method.
Our key contributions are summarized as follows:

\begin{itemize}
    \item We first recognize the crucial role that gradient distribution shifts play in noisy label learning and propose a simple yet effective strategy to obtain a dynamic threshold that adapts to these continual shifts.
    \item We provide a theoretical analysis demonstrating that the threshold obtained by OGC can make non-robust loss functions robust against various types of label noise.
    \item We conduct extensive experiments to evaluate the effectiveness of our proposed method across multiple types of label noise, including symmetric, asymmetric, instance-dependent, and real-world noise.
\end{itemize}

\section{Related Works}

\textbf{Robust loss functions.}
In recent years, several robust-loss-based methods have been developed for learning with noisy labels. 
\citet{DBLP:conf/aaai/GhoshKS17:RobustLoss} theoretically demonstrated the robustness of symmetric loss functions, such as Mean Absolute Error (MAE), to label noise.
Building on this, \citet{DBLP:conf/nips/ZhangS18:GCE} introduced the Generalized Cross Entropy (GCE), which is a generalization of Cross Entropy (CE) and MAE.
Further, \citet{DBLP:conf/iccv/0001MCLY019:SCE} combined CE with scaled MAE to propose Symmetric Cross Entropy (SCE).
Additionally, \citet{DBLP:conf/iclr/MenonRRK20:PHuberCE} introduced a composite loss-based gradient clipping applied to CE, resulting in PHuber-CE.
\citet{DBLP:conf/ijcai/FengSLL0020:Taylor-CE} utilized the Taylor series to derive an alternative representation of CE, leading to the development of Taylor-CE.
Moreover, \citet{DBLP:conf/icml/MaH00E020:NormalizedLoss} proposed the Active Passive Loss (APL), aiming to create fully robust loss functions.
\citet{DBLP:conf/icml/ZhouLJGJ21:AsymmetricLoss} developed Asymmetric Loss Functions (ALFs) to address limitations posed by symmetric conditions.
\citet{DBLP:conf/icml/WeiZXF00L23:LogitClip} introduced logit clipping (LogitClip), a technique that clamps the norm of the logit vector to a constant upper bound.
Recently, \citet{DBLP:conf/nips/YeLD0ST23:NormalizedNegativeLoss} proposed Normalized Negative Loss Functions (NNLFs), which prioritize learning from well-understood samples.

\textbf{Other methods for noisy label learning.}
In addition to robust loss functions, various other approaches are utilized for learning with noisy labels.
A sample selection-based approach aims to identify and select clean samples for training, often leveraging the disagreement between two models \cite{DBLP:conf/icml/Yu0YNTS19:Co-teaching+, DBLP:conf/cvpr/WeiFC020:JoCoR}.
Some research further combines these methods with techniques from unsupervised or semi-supervised learning to make full use of the unselected corrupted samples \cite{DBLP:conf/iclr/NguyenMNNBB20:SELF, DBLP:conf/iclr/LiSH20:DivideMix, DBLP:conf/eccv/WeiSLY22:SelfFiltering}.
A loss correction-based approach modifies the loss of each sample based on label-dependent weights \cite{DBLP:conf/nips/NatarajanDRT13:LNL}, an estimated noise transition matrix \cite{DBLP:conf/nips/HanYNZTZS18:Masking, DBLP:conf/cvpr/PatriniRMNQ17:LossCorrection}, or model predictions \cite{DBLP:conf/icml/ArazoOAOM19:UnsupLossCorr}.
Additionally, a regularization-based approach is also employed to combat label noise, including methods such as MixUp \cite{DBLP:conf/iclr/ZhangCDL18:MixUp}, label smoothing \cite{DBLP:conf/icml/LukasikBMK20:LabelSmooth}, or leveraging previous prediction results \cite{DBLP:conf/nips/LiuNRF20:ELR}.

\section{Preliminaries}

\subsection{Risk Minimization and Noisy Labels}
\label{sec:risk-noise}

In this work, we consider a typical classification problem with \(K\)-categories.
Let $\mathcal{X} \subset \mathbb{R}^d$ be the sample space and $\mathcal{Y} = [K] \defeq \{1, \cdots, K\}$ be the label space.
Given a dataset \(\{(\boldsymbol{x}_n, y_n)\}_{n=1}^N\), where each sample pair \((\boldsymbol{x}_n, y_n)\) is drawn \emph{i.i.d.} from an underlying ``clean'' distribution $\mathcal{D}_c$, over $\mathcal{X} \times \mathcal{Y}$.
We define \(\boldsymbol{q}(k|\boldsymbol{x}_n)\) as the ground truth probabilistic distribution across various labels for a sample \(\boldsymbol{x}_n\), along with the condition that \(\sum_{k=1}^K \boldsymbol{q}(k|\boldsymbol{x}_n) = 1\).
Specifically, \(\boldsymbol{q}(k|\boldsymbol{x}_n)\) is a degenerate distribution where \(\boldsymbol{q}(y_n|\boldsymbol{x}_n) = 1\) and \(\boldsymbol{q}(k \ne y_n|\boldsymbol{x}_n) = 0\).

We define a classifier $f: \mathcal{X} \to \mathbb{R}^K$ as a function that maps from sample space to logit space, parameterized by trainable parameters $\theta$.
Specifically, we focus on the case that $f$ is a deep neural network (DNN).
Given a sample $\boldsymbol{x}_n$, the function $f(\boldsymbol{x}_n)$ computes the logits.
By applying a softmax output layer, we obtain the \textbf{model-predicted probability} $\boldsymbol{p}(k|\boldsymbol{x}_n)$ for each category $k \in \{1,\cdots,K\}$, with $\sum_{k=1}^K \boldsymbol{p}(k|\boldsymbol{x}_n)=1$.
Training a classifier $f$ is to find an optimal classifier $f^*$ that minimize the empirical risk: $\sum_{n=1}^N \ell(f(\boldsymbol{x}_n), y_n)$, where $\ell: \mathbb{R}^K \times \mathcal{Y} \to \mathbb{R}^+$ denotes the loss function.

When label noise exists, we work with a corrupted dataset \(\{(\boldsymbol{x}_n, \tilde{y}_n)\}_{n=1}^N\), where each sample is drawn \emph{i.i.d.}\ from an unknown distribution \(\mathcal{D}_\eta\) and \(\tilde{y}\) denotes the possibly incorrect label.
There are two ways to model label noise: instance-independent or instance-dependent.

In the \textbf{instance-independent} approach, it is assumed that, given the true label $y$, the corruption process is conditionally independent of input sample $\boldsymbol{x}$ \cite{DBLP:conf/aaai/GhoshKS17:RobustLoss}.
So the stochastic corruption that the true label \(y\) is corrupted into label \(j\) is specified by a transition matrix,
\begin{equation}
\label{eqn:ins-ind}
    [\eta_{yj}] = q_\eta(j|y), \qquad \forall y, j \in [K],
\end{equation}
where $\eta_{yj}$ denotes the probability that true label $y$ is corrupted into label $j$, and $\eta_y=\sum_{j \ne y}\eta_{yj}$ denotes the noise rate for label $y$.
Under this model assumption, label noise can be either symmetric or asymmetric.
For symmetric noise, $\eta_{ij}=\frac{\eta_i}{K-1}, \forall j \ne i$ and $\eta_i = \eta, \forall i \in [k]$, where $\eta$ is a constant.
For asymmetric noise, $\eta_{ij}$ varies depending on both the true label $i$ and corrupted label $j$.

The \textbf{instance-dependent} approach extends the setting above and assumes that the corruption process is conditionally dependent on both the true label \(y\) and input sample \(\boldsymbol{x}\) \cite{DBLP:conf/nips/XiaL0WGL0TS20:PDN}.
Similarly, the probability of the event that true label \(y\) is corrupted into label \(j\) is given by:
\begin{equation}
\label{eqn:ins-dep}
    \eta_{yj}(\bm x) = q_\eta(j|y, \boldsymbol{x}),
\end{equation}
where \(\eta_{yj}(\bm x)\) now depends on $\bm x$, and $\eta_y(\bm x) = \sum_{j \ne y} \eta_{yj}(\bm x)$ dentes the noise rate for sample $\bm x$.

As a closing remark, the unknown ``noisy'' distribution \(\mathcal{D}_\eta\) can be taken as a mixture of the ``clean'' distribution $\m D_c$ and the ``noise'' distribution $\m D_n$.
Specifically, $q_\eta(\bm x, \tilde y) = q_\eta(\tilde y = y) \cdot q_\eta(\bm x, \tilde y \mid \tilde y = y) + q_\eta(\tilde y \neq y) \cdot q_\eta(\bm x, \tilde y \mid \tilde y \neq y)$;
$q_\eta(\bm x, \tilde y \mid \tilde y = y)$ is exactly the previously introduced ``clean'' distribution $\m D_c$, and we further denote $\m D_n \vcentcolon= q_\eta(\bm x, \tilde y \mid \tilde y \neq y)$.

\subsection{Gradient Clipping}

We first introduce the vector clipping function. Given a user-specified threshold \(\tau > 0\), the clipping function can be formally defined by:
\begin{equation}
    \texttt{clip}(w,\tau) = \begin{cases}
        \tau \cdot \frac{w}{\|w\|_2}, & \text{if} \quad \|w\|_2 \ge \tau, \\
        w, & \text{otherwise},
    \end{cases}
\end{equation}
where \(w\) denotes the vector to be clipped, and the output is guaranteed to have an $\ell_2$ norm upper bounded by $\tau$.

In practice, the clipping function is usually applied to a mini-batch \(\{(x_m, y_m)\}_{m=1}^M\).
To clip the gradients for the model parameters \(\theta\), the operation is defined as follows:
\begin{align}
    & \qquad \bar{g}_\theta = \texttt{clip}(g_\theta, \tau), \nonumber \\
    & \text{where} \quad g_\theta = \frac{1}{M} \sum_{m=1}^M \nabla_\theta \ell(f(x_m), y_m).
\end{align}
Standard gradient clipping is typically employed to avoid gradient explosion \cite{DBLP:conf/icml/PascanuMB13:GradientExplosion} and accelerate the convergence of the model \cite{DBLP:conf/iclr/ZhangHSJ20:GradientCLipAccelTrain}.

As an implicit regularization approach, gradient clipping is adopted by practitioners to alleviate the issues caused by noisy labels.
However, \citet{DBLP:conf/iclr/MenonRRK20:PHuberCE} suggested that merely clipping the gradient of the model parameter \( \theta \) does not effectively counteract the detrimental impacts of noisy labels.
They discovered that, on a linear model, applying gradient clipping to parameters is equivalent to applying it to logits.
However, clipping gradients on logits modifies the loss function in a way that allows outliers to still influence the gradients, ultimately compromising the robustness.
Instead, they recommend applying gradient clipping directly to the gradients of the model predicted probabilities \(\bm p\) in the following manner:
\begin{equation}
    \bar{g}_{\bm p} = \texttt{clip}(g_{\bm p},\tau), \quad \text{where} \quad g_{\bm p} = \nabla_{\bm p} \ell(f(\boldsymbol{x}), y).
\end{equation}
Applying gradient clipping to probabilities results in a modified loss function where the loss value remains nearly constant when logits exceed a certain threshold.
This property helps the loss function assign small gradients to outlier samples (which are likely to be corrupted).
Consequently, it bounds the gradient caused by noisy samples, enhancing the robustness of the loss function against noisy labels.

\section{Optimized Gradient Clipping}
\label{sec: Optimized Gradient Clipping}

We present the methodology of optimized gradient clipping in this section.

\subsection{Motivation}

Given the observation that a fixed clipping threshold may be ineffective in addressing the inherent variability of gradient distributions, we reconsider the problem of selecting the clipping threshold from a new perspective.

Specifically, we are interested in understanding how the clipping threshold \(\tau^{(t)}\) affects \(r^{(t)}\), which represents the ratio of expected gradients between clean and noisy samples:
\begin{equation}
    r^{(t)} = \frac{\mathbb{E}_{(\bm x, \tilde{y}) \sim \mathcal{D}_n} [ \|\texttt{clip}(\nabla_{\bm p} \ell(f(\bm x), \tilde{y}), \tau^{(t)})\|_2 ]}{\mathbb{E}_{(\bm x, \tilde y) \sim \mathcal{D}_c}[\|\texttt{clip}(\nabla_{\bm p} \ell(f(\bm x), \tilde y), \tau^{(t)})\|_2]}.
    \label{true ratio}
\end{equation}
This formula calculates the ratio of expected gradient (with respect to the predicted probabilities) between noise and clean samples after clipping at time step \(t\).
Intuitively, this ratio reflects the extent to which noisy samples influence the training process.
As the ratio increases, the influence of noisy samples begins to dominate, leading to model overfitting.
Conversely, by controlling this ratio through the optimized  threshold $\tau^{(t)}$, we can effectively manage the impact of noisy samples.

\subsection{Method}

Here, we formally introduce our method, named as Optimized Gradient Clipping (OGC).
Our approach consists of the following three components:
1) Modeling clean and noisy distributions,
2) Determining the clipping threshold, and
3) Applying clipping to the loss functions.
Our algorithm is outlined in Appendix.

To simplify our discussion, we focuses on a specific type of loss function under the following assumption: the loss value $\ell$ depends exclusively on \( p(y|x) \), the predicted probability of the given label.
It should be noted that most widely used loss functions follow this assumption.
This includes both non-robust loss functions, such as Cross Entropy (CE) and Focal Loss (FL), and robust loss functions, such as Generalized Cross Entropy (GCE).

This assumption enables us to define two key mapping functions:
1) $\phi_{H \to \ell}$, which directly maps the cross entropy value $H$ to the loss value $\ell$; and
2) $\phi_{\|g_{\bm p}\|_2 \to \ell}$, which directly maps $\|g_{\bm p}\|_2$, the gradient norm of the predicted probability,  to the loss value $\ell$.

\textbf{Modeling of clean and noisy distributions.}
In the real world, we only have access to a corrupted dataset sampled from an unknown distribution \(D_\eta\), making the direct calculation of \(r^{(t)}\) unfeasible.
To address this, we propose a proxy ratio \(\tilde{r}^{(t)}\) to approximate the ground-truth \(r^{(t)}\).

Given a mini-batch \(\{(\boldsymbol{x}_m, \tilde{y}_m)\}_{m=1}^M\) sampled at time \(t\), we recall the cross entropy for a sample pair \((\boldsymbol{x}, \tilde{y})\) is given as follows:
\begin{align}
    H(f(\boldsymbol{x}), \tilde{y}) &= - \sum_{k=1}^K q(k|\boldsymbol{x}) \log p(k|\boldsymbol{x}) \nonumber \\
    &= - \log p(\tilde{y}|\boldsymbol{x}).
\end{align}
By respectively assuming $(\boldsymbol{x}, \tilde{y}) \sim \m D_c$ and $(\boldsymbol{x}, \tilde{y}) \sim \m D_n$, $H(f(\boldsymbol{x}), \tilde{y})$ will exhibit two distinct distributions,
which accordingly induce two random variables $H_c, H_n$.

The two-component Gaussian Mixture Model (2-GMM), which can be considered a non-parametric method, is commonly utilized to model the Cross Entropy value \(H\) in noisy label learning scenarios \cite{DBLP:conf/iclr/LiSH20:DivideMix}.
This model enabling us to effectively characterize the clean and noisy distributions.
Here, we fit a 2-GMM on the empirical distribution of $H(f(\boldsymbol{x}), \tilde{y})$ for $(\boldsymbol{x}, \tilde{y}) \sim \m D_\eta$, resulting in the following two Gaussian approximation distributions:
\begin{gather}
    H_c \sim \mathcal{N}_c^{(t)}, \quad \text{where} \quad \mathcal{N}_c^{(t)} = \mathcal{N}(\mu_c^{(t)}, {\sigma_c^{(t)}}^2) \\
    H_n \sim \mathcal{N}_n^{(t)}, \quad \text{where} \quad \mathcal{N}_n^{(t)} = \mathcal{N}(\mu_n^{(t)}, {\sigma_n^{(t)}}^2),
\end{gather}
where \(\mu_c^{(t)} < \mu_n^{(t)}\), \(\mathcal{N}_c^{(t)}\) and \(\mathcal{N}_n^{(t)}\) denotes the \texttt{c}lean and \texttt{n}oise gaussian distribution at time \(t\), respectively.
It should be noted that the assumption that clean samples exhibit lower cross entropy values than noisy samples is widely accepted in the field of noisy label learning \cite{DBLP:conf/ijcai/GuiWT21:SmallLossCriterion}.

\textbf{Determine clipping threshold \(\tau^{(t)}\).}
Given the clean and noise distributions, we can calculate \(\tilde{r}^{(t)}\) in the following manner
($\phi_{H \to \ell}(\cdot)$ is defined in the third paragraph of this subsection):
\begin{equation}
    \tilde{r}^{(t)}
    = \frac{\int_0^{+\infty} \| \texttt{clip}(\nabla_p \phi_{H \to \ell}(H), \tau^{(t)}) \|_2 \mathrm{d} \mathbb{P}_n (H)}{\int_0^{+ \infty} \| \texttt{clip}(\nabla_p \phi_{H \to \ell}(H), \tau^{(t)}) \|_2 \mathrm{d} \mathbb{P}_c (H)},
\end{equation}
where \(\mathbb{P}_c\) and \(\mathbb{P}_n\) are the distribution functions of the clean and noise Gaussian distributions \(\mathcal{N}_c^{(t)}\) and \(\mathcal{N}_n^{(t)}\), respectively.
And the clipping threshold \(\tau^{(t)}\) can be obtained by solving the following optimizing problem:
\begin{equation}
    \min_{\tau^{(t)}} \tilde{r}^{(t)}(\tau^{(t)}).
    \label{optimize r original}
\end{equation}
However, directly optimizing Eq.\ref{optimize r original} may result in excessive clipping.
Specifically, one could clip as many gradients as possible to minimize the ratio.
In such cases, although most of the noisy gradients are effectively removed, the clean gradients are also impacted, leading to model underfitting.
To address this, we instead employ a conditioned version of Eq.\ref{optimize r original} to balance the clipping of noisy gradients while preserving the clean gradients:
\begin{equation}
    \begin{aligned}
        \min_{\tau^{(t)}} \tilde{r}^{(t)}(\tau^{(t)}) \quad \text{s.t.} \quad \tilde{r}^{(t)} \geq 1 + \epsilon,
    \end{aligned}
    \label{optimize to get tau_t}
\end{equation}
where \(\epsilon > 0\) is a hyper-parameter, and $1 + \epsilon$ is the predefined gradient ratio.
This allows us to consistently determine a clipping threshold \(\tau^{(t)}\) that preserves the desired ratio, regardless of variations in the gradient distributions of clean and noisy samples, while preventing excessive clipping of the clean gradients.
In practice, a straightforward binary search algorithm is applied to solve Eq.\ref{optimize to get tau_t}.

\textbf{Implement optimized clipping on loss functions.}
According to \citet[Lemma 5]{DBLP:conf/iclr/MenonRRK20:PHuberCE}, a loss function \(\ell\) equipped with gradient clipping is equivalent to a Huberised-like loss function as follows:
if \( \|\nabla_{\bm{p}}(\ell(f(\bm{x}), y))\|_2 \geq \tau^{(t)}\), then \(\bar{\ell}(f(\bm{x}), y, \tau^{(t)}) = 1 - \tau^{(t)} \cdot p(y|\bm{x}) + \phi_{\|g_{\bm{p}}\|_2 \to \ell}(\tau^{(t)})\);
otherwise, \(\bar{\ell}(f(\bm{x}), y, \tau^{(t)}) = \ell(f(\bm{x}), y)\).

Take Cross Entropy (CE) loss \(\ell_\text{CE}\) as an example.
By applying Optimized Gradient Clipping (OGC), we obtain CE+OGC, which is induced by a new loss function \(\bar{\ell}_\text{CE}(f(\bm{x}), y, \tau^{(t)})\) as follows:
if \( \frac{1}{p(y|\bm{x})} \geq \tau^{(t)}\), then \(\bar{\ell}_\text{CE}(f(\bm{x}), y, \tau^{(t)}) = 1 - \tau^{(t)} \cdot p(y|\bm{x}) + \log \tau^{(t)}\);
otherwise, \(\bar{\ell}_\text{CE}(f(\bm{x}), y, \tau^{(t)}) = -\log p(y|\bm{x})\).

\begin{table*}[htbp]
    \centering
    \small
    \begin{tabular}{ccccccc}
        \toprule
        Methods & Sym-50\% & Sym-80\% & Asymmetric & Dependent & Real & Average \\
        \midrule
        Fixed & 84.56 & 34.21 & 58.95 & 44.24 & 79.84 & 60.36 \\
        Linear & 81.01 & 42.01 & 77.90 & 65.68 & 75.63 & 68.44 \\
        EMA & 84.98 & 21.31 & 81.30 & 73.06 & \textbf{82.23} & 68.57 \\
        Optimized (ours) & \textbf{85.16} & \textbf{43.77} & \textbf{81.51} & \textbf{78.28} & 81.45 & \textbf{74.03} \\
        \bottomrule
    \end{tabular}
    \caption{
        Test accuracies (\%) of \(\tau^{(t)}\) adjusting methods on CIFAR-10 dataset with different label noise,
        \textbf{bold} for best results.
    }
    \label{table: adjusting methods}
\end{table*}

\subsection{Robustness Analysis}

In the following, we provide a formal analysis of the robustness exhibited by our proposed OGC.
Specifically, we will first introduce an indicator of robustness, ``excess risk'', which is the focus of our analysis. 
To simplify our discussion, we specifically analyze the new loss function $\bar{\ell}_\text{CE}$ induced by CE+OGC, which is Cross Entropy equipped with OGC.
(Our results can be easily generalized to other loss functions, such as Focal Loss (FL) and Generalized Cross Entropy (GCE).)

Given any classifier $f$ and loss function $\bar{\ell}_\text{CE}$ with $\tau^{(t)}$, the population risks of $f$ under the clean distribution $\mathcal{D}_c$ and the unknown distribution $\mathcal{D}_\eta$ are respectively defined as
\begin{align}
    \mathcal{R}_{\bar{\ell}_\text{CE}}(f, \tau^{(t)}) = \mathbb{E}_{(\bm{x}, y) \sim \mathcal{D}_c}[\bar{\ell}_\text{CE}(f(\bm{x}), y, \tau^{(t)})],
    \nonumber \\
    \mathcal{R}^\eta_{\bar{\ell}_\text{CE}}(f, \tau^{(t)}) = \mathbb{E}_{(\bm{x}, \tilde{y}) \sim \mathcal{D}_\eta}[ \bar{\ell}_\text{CE}( f(\bm{x}), \tilde{y}, \tau^{(t)})].
\end{align}
Let $f^\star$ be the global minimizer of $\mathcal{R}_{\bar{\ell}_\text{CE}}(f, \tau^{(t)})$, and let $\tilde{f}^\star$ be the global minimizer of $\mathcal{R}^\eta_{\bar{\ell}_\text{CE}}(f, \tau^{(t)})$.
With the notations, we can then define the ``excess risk'' as $\mathcal{R}_{\bar{\ell}_\text{CE}}(\tilde{f}^\star, \tau^{(t)}) - \mathcal{R}_{\bar{\ell}_\text{CE}}(f^\star, \tau^{(t)})$ or $\mathcal{R}^\eta_{\bar{\ell}_\text{CE}}(f^\star, \tau^{(t)}) - \mathcal{R}^\eta_{\bar{\ell}_\text{CE}}(\tilde{f}^\star, \tau^{(t)})$ (which indicates the performance gap of $\tilde{f}^\star$ and $f^\star$).

We first show the boundedness of $\bar{\ell}_\text{CE}$, which is important to the control of ``excess risk''.

\begin{proposition}
    \label{proposition: bounded loss}
    Given any classifier \(f\), for any input sample pair \((\bm x, y)\) and any \(\tau^{(t)} \ge 1\), the CE+OGC loss $\bar{\ell}_\text{CE}$ is both lower and upper bounded:
    \begin{align}
        1 - p(j|x) &\le \bar{\ell}_\text{CE}(f(\bm x), j, \tau^{(t)}) \nonumber \\
        &\le (1 - p(j|\bm x)) (1 + \log \tau^{(t)}).
    \end{align}
    Moreover, the sum of $\bar{\ell}_\text{CE}$ w.r.t\ all classes is thus also lower bound and upper bounded:
    \begin{align}
        K - 1 &\le \sum_{j=1}^K \bar{\ell}_\text{CE}(f(\bm x), j, \tau^{(t)}) \nonumber \\
        &\le (K - 1) (1 + \log \tau^{(t)}).
    \end{align}
\end{proposition}

The proof is deffered to Appendix. 
Proposition~\ref{proposition: bounded loss} demonstrates that Cross Entropy equipped with our proposed OGC is always bounded.
Moreover, as $\tau^{(t)}$ approaches 1, the bound becomes tighter.
When $\tau^{(t)} = 1$, our loss function is equivalent to the symmetric Mean Absolute Error (MAE).
Notably, this symmetry is a key property that makes a loss function robust to label noise \cite{DBLP:conf/aaai/GhoshKS17:RobustLoss}.

Proposition~\ref{proposition: bounded loss} paves the way for further noise robustness analysis of CE+OGC loss $\bar{\ell}_\text{CE}$.
Firstly, under the common instance-independent setting, we can depict the bound of excess risk for both symmetric and asymmetric label noise (see Equation~(\ref{eqn:ins-ind}) for the definitions).

\begin{theorem}[Excess risk under instance-independent symmetric label noise]
    \label{theorem: noise-tolerant under symmetric noise}
    Under symmetric label noise with $\eta \le 1 - \frac{1}{K}$,
    \begin{align}
        0 &\le \mathcal{R}_{\bar{\ell}_\text{CE}}(\tilde{f}^\star, \tau^{(t)}) - \mathcal{R}_{\bar{\ell}_\text{CE}}(f^\star, \tau^{(t)}) \nonumber \\
        &\le \log \tau^{(t)} \Big/ \Big( 1 - \frac{\eta K}{K - 1} \Big).
    \end{align}
\end{theorem}

\begin{table*}[htbp]
\centering
\small
\begin{tabular}{ccccccc}
    \toprule
    Methods & Sym-50\% & Sym-80\% & Asymmetric & Dependent & Real & Average\\
    \midrule
    GCE & 85.36 & 34.72 & 59.87 & 66.76 & 81.33 & 65.61\\
    GCE+OGC & 86.23 & 38.68 & 80.88 & 77.22 & 81.75 & 72.95\\
    Improvement & + 0.87 & + 3.96 & + 21.01 & + 10.46 & + 0.42 & + 7.34\\
    \bottomrule
\end{tabular}
\caption{Test accuracies (\%) of GCE and GCE+OGC on CIFAR-10 dataset with different label noise.}
\label{table: GCE+OGC}
\end{table*}

\begin{table*}[t]
\small
\centering
\begin{tabular}{cccccccc}
    \toprule
    Methods & Sym-20\% & Sym-50\% & Sym-80\% & Asymmetric & Dependent & Real & Average\\
    \midrule
    CE & 81.25$\pm$0.35 & 51.58$\pm$0.41 & 35.97$\pm$5.46 & 76.20$\pm$0.38 & 60.37\(\pm\)0.68 & 64.43\(\pm\)0.24 & 61.63 \\
    FL & 82.32$\pm$0.39 & 52.65$\pm$0.35 & 37.30$\pm$1.13 & 76.69$\pm$0.15 & 61.12$\pm$0.31 & 65.57$\pm$0.39 & 62.61\\
    MAE & 89.99$\pm$0.14 & 75.80$\pm$3.72 & 18.81$\pm$2.01 & 56.09$\pm$0.29 & 15.70\(\pm\)1.79 & 55.93\(\pm\)3.67 & 52.05 \\
    GCE & 91.24\(\pm\)0.13 & \underline{85.36\(\pm\)0.20} & 34.72$\pm$3.65 & 59.87\(\pm\)0.10 & 66.76$\pm$0.38 & 81.33\(\pm\)0.06 & 69.88 \\
    SCE & 91.34\(\pm\)0.05 & 84.91\(\pm\)0.33 & 40.59\(\pm\)0.48 & 79.52\(\pm\)0.41 & 77.75\(\pm\)1.08 & 81.20\(\pm\)0.15 & 75.89 \\
    PHuber-CE & 90.75\(\pm\)0.13 & 84.56\(\pm\)0.14 & 34.21\(\pm\)3.91 & 58.95\(\pm\)0.23 & 44.24\(\pm\)10.21 & 79.84\(\pm\)0.17& 65.43 \\
    Taylor-CE & 91.24\(\pm\)0.08 & 84.60\(\pm\)0.10 & \underline{43.69\(\pm\)5.26} & 59.19\(\pm\)0.23 & 57.26\(\pm\)6.73 & 80.08\(\pm\)0.14 & 69.34 \\
    NCE+RCE & 91.21\(\pm\)0.09 & 84.95\(\pm\)0.27 & 26.38\(\pm\)3.05 & 78.58\(\pm\)0.32 & \underline{78.75\(\pm\)0.37} & 80.78\(\pm\)0.21 & 73.44\\
    JS & 91.65\(\pm\)0.05 & 83.79\(\pm\)0.21 & 40.74\(\pm\)5.98 & 75.45\(\pm\)0.35 & 68.24$\pm$1.10 & 78.55$\pm$0.31 & 71.40\\
    LC-CE & \underline{91.69\(\pm\)0.08} & 79.29\(\pm\)0.35 & 35.77\(\pm\)3.27 & 75.48\(\pm\)0.06 & 61.19$\pm$0.40 & 74.67$\pm$0.43 & 69.68\\
    NCE+NNCE & 90.96\(\pm\)0.08 & 85.19\(\pm\)0.28 & 17.56\(\pm\)0.36 & 77.63\(\pm\)0.26 & \textbf{80.92$\pm$0.66} & \underline{82.39$\pm$0.08} &72.44\\
    \midrule
    CE+OGC & \textbf{91.80$\pm$0.13} & 85.16$\pm$0.26 & \textbf{43.77$\pm$2.27} & \underline{81.51$\pm$0.31} & 78.28$\pm$0.12 & 81.45$\pm$0.68 & \underline{76.99}\\
    FL+OGC & 88.51$\pm$0.41 & \textbf{85.43$\pm$0.13} & 41.43$\pm$1.02 & \textbf{86.34$\pm$0.49} & 78.21$\pm$0.86 & \textbf{82.69$\pm$0.30} & \textbf{77.10}\\
    \bottomrule
\end{tabular}
\caption{
    Test accuracies (\%) of different methods on CIFAR-10 datasets with different label noise.
    The results (mean$\pm$std) are reported over 3 random runs.
    \textbf{Bold} denotes the best results and \underline{underline} denotes the second-best results.}
\label{table: cifar10}
\end{table*}

\begin{table*}[t]
\small
\centering
\begin{tabular}{cccccccc}
    \toprule
    Methods & Sym-20\% & Sym-50\% & Sym-80\% & Asymmetric & Dependent & Real & Average\\
    \midrule
    CE & 64.75\(\pm\)0.22 & 49.56\(\pm\)0.63 & 8.94\(\pm\)0.53 & 45.32\(\pm\)0.23 & 48.78\(\pm\)0.33 & 54.64\(\pm\)0.09 & 45.33\\
    FL & 64.58$\pm$0.77 & 50.27$\pm$0.28 & 9.52$\pm$0.55 & 46.55$\pm$0.14 & 49.67$\pm$0.35 & 53.80$\pm$0.11 & 45.73\\
    MAE & 5.46\(\pm\)0.69 & 3.63\(\pm\)0.12 & 0.99\(\pm\)0.01 & 2.81\(\pm\)0.45 & 1.36$\pm$0.27 & 2.79$\pm$0.80 &14.86\\
    GCE & 71.14\(\pm\)0.23 & 65.18\(\pm\)0.52 & 11.99\(\pm\)0.37 & 42.59\(\pm\)0.56 & 51.43$\pm$0.69 & 56.58$\pm$0.21 & 49.65\\
    SCE & 66.01\(\pm\)0.23 & 57.00\(\pm\)0.37 & 6.44\(\pm\)0.17 & 39.93\(\pm\)0.41 & 45.53$\pm$0.46 & 53.45$\pm$0.13 & 45.23\\
    PHuber-CE & 54.08\(\pm\)1.67 & 48.94\(\pm\)0.34 & 15.79\(\pm\)1.07 & 31.97\(\pm\)0.41 & 20.22$\pm$0.91 & 34.71$\pm$0.66 & 37.33\\
    Taylor-CE & 69.87\(\pm\)0.36 & 63.84\(\pm\)0.12 & 23.10\(\pm\)0.59 & 41.94\(\pm\)0.29 & 43.44$\pm$0.75 & 49.76$\pm$0.48 & 50.20\\
    NCE+RCE & 69.95\(\pm\)0.19 & 61.31\(\pm\)0.10 & 14.87\(\pm\)0.39 & 42.65\(\pm\)0.25 & 51.24$\pm$0.22 & 56.40$\pm$0.30 & 48.74\\
    JS & 71.15\(\pm\)0.34 & 65.60\(\pm\)0.11 & 17.15\(\pm\)0.37 & 44.82\(\pm\)0.60 & 51.37$\pm$0.87 & 55.17$\pm$1.20 & 51.07\\
    LC-CE & \underline{71.40\(\pm\)0.16} & 62.61\(\pm\)0.11 & 13.16\(\pm\)0.54 & 42.84\(\pm\)0.25 & 47.01$\pm$0.52 & 55.87$\pm$0.35 & 48.65\\
    NCE+NNCE & 69.07\(\pm\)0.24 & 62.09\(\pm\)0.12 & 7.93\(\pm\)0.88 & \textbf{50.86\(\pm\)0.25} & 55.31$\pm$2.13 & 57.55$\pm$0.19 & 50.81\\
    \midrule
    CE+OGC & \textbf{71.41$\pm$0.42} & \textbf{67.61$\pm$0.46} & \underline{28.70$\pm$1.54} & 45.37$\pm$0.25 & \underline{57.28$\pm$1.40} & \underline{59.93$\pm$0.15} & \underline{54.78}\\
    FL+OGC & 71.13$\pm$0.34 & \underline{66.65$\pm$0.09} & \textbf{31.33$\pm$0.42} & \underline{47.84$\pm$1.06} & \textbf{60.83$\pm$0.85} & \textbf{60.30$\pm$0.28} & \textbf{56.36}\\
    \bottomrule
\end{tabular}
\caption{
    Test accuracies (\%) of different methods on CIFAR-100 datasets with different label noise.
    The results (mean$\pm$std) are reported over 3 random runs.
    \textbf{Bold} denotes the best results and \underline{underline} denotes the second-best results.}
\label{table: cifar100}
\end{table*}

\begin{theorem}[Excess risk under instance-independent asymmetric label noise]
    \label{theorem: noise-tolerant under asymmetric noise}
    Under asymmetric label noise with $\eta_{ij} < 1 - \eta_i, \forall j \ne i, \forall i, j \in [k]$, where $1 - \eta_i = \eta_{ii}$, if $\mathcal{R}_{\bar{\ell}_\text{CE}}(f^\star, \tau^{(t)}) = 0$, then
    \begin{align}
        0 & \le \mathcal{R}^\eta_{\bar{\ell}_\text{CE}}(f^\star, \tau^{(t)}) - \mathcal{R}^\eta_{\bar{\ell}_\text{CE}}(\tilde{f}^\star, \tau^{(t)}) \nonumber \\
        & \le (K - 1)(\log \tau^{(t)}) \mathbb{E}_{(\bm x, y) \sim D_c} [1 - \eta_y].
    \end{align}
\end{theorem}

The proofs for Theorem~\ref{theorem: noise-tolerant under symmetric noise} and Theorem~\ref{theorem: noise-tolerant under asymmetric noise} can be found in Appendix.
These theorems illustrate that, under both symmetric and asymmetric label noise, with our proposed OGC method, the excess risk between $\tilde{f}^\star$ and $f^\star$ is consistently bounded. Meanwhile, these bounds get tighter as $\tau^{(t)}$ decreases.

Next, we further extend our analysis to scenarios where the label noise is instance-dependent (see Equation~\ref{eqn:ins-dep}).
With a regular mild assumption on signal-to-noise ratio that $\eta_{yj}(\bm x) < \eta_{yy}(\bm x), \forall \bm x, j \ne y$, we have the following result:

\begin{theorem}[Excess risk under instance-dependent label noise]
    \label{theorem: noise-tolerant under instance-dependent noise}
    Under instance-dependent label noise with $\eta_{yj}(\bm x) < \eta_{yy}(\bm x), \forall \bm x, j \ne y$, if $\mathcal{R}_{\bar{\ell}_\text{CE}}(f^\star, \tau^{(t)}) = 0$, then
    \begin{align}
        0
        &\le \mathcal{R}^\eta_{\bar{\ell}_\text{CE}}(f^\star, \tau^{(t)}) - \mathcal{R}^\eta_{\bar{\ell}_\text{CE}}(\tilde{f}^\star, \tau^{(t)}) \nonumber \\
        &\le (K - 1)(\log \tau^{(t)}) \mathbb{E}_{(\bm x, y) \sim D_c} [1 - \eta_y(\bm x)].
    \end{align}
\end{theorem}

The proof for Theorem~\ref{theorem: noise-tolerant under instance-dependent noise} is provided in Appendix.
This theorem demonstrates that, even under instance-dependent label noise, our proposed OGC method ensures that the risk differences between $\tilde{f}^\star$ and $f^\star$ remain consistently bounded.

\begin{table*}[t]
\centering
\small
\begin{tabular}{@{}c@{\hspace{1em}}cccccccc@{}}
    \toprule
    Method & CE & GCE & SCE & PHuber-CE & NCE+RCE & LC-CE & CE+OGC & FL+OGC \\
    \midrule
    ILSVRC2012 Val & 64.96 & 56.60 & 63.47 & 63.75 & 65.08 & 66.47 & \underline{67.40} & \textbf{68.28} \\
    WebVision Val & 68.51 & 58.20 & 65.63 & 65.43 & 66.75 & 69.56 & \underline{70.28} & \textbf{70.60} \\
    \bottomrule
\end{tabular}
\caption{
    Top-1 validation accuracies (\%) of different methods on the ILSVRC12 and WebVision validation sets, under the ``Mini'' setting \cite{DBLP:conf/icml/JiangZLLF18:MentorNet}.
    \textbf{Bold} denotes the best results and \underline{underline} denotes the second-best results.
}
\label{table: webvision}
\end{table*}

\section{Experiments}
\label{sec: experiments}

\subsection{Empirical Understanding}
\label{sec: empirical understanding}

In this subsection, to enhance our understanding of the proposed OGC, we conduct a series of experiments.
Unless otherwise stated, our experimental settings follow those described in Section~\ref{CIFAR-10 and CIFAR-100}.
For additional experimental results on parameter analysis, including predefined gradient ratio $1 + \epsilon$, queue \( Q \), time frame \( s \), and time consumption, please refer to Appendix.

\textbf{Comparison with manually designed methods.}
To highlight the advantages of our optimize-based strategy, we compare it on CIFAR-10 against manually designed approaches for tuning \(\tau^{(t)}\):
1) Linear Decrease Strategy: This approach is defined by \(\tau^{(t)} = \beta \cdot ( 1 - \frac{t}{T} ) \), where \(T\) represents the total number of time steps and \(\beta\) is a parameter.
2) Exponential Moving Average (EMA) Strategy: This approach uses the formula \(\frac{1}{\tau^{(t)}} = \alpha \cdot \frac{1}{\tau^{(t-1)}} + (1 - \alpha) \cdot 1\), where \(\alpha\) is a parameter.
We also compare our optimize-based strategy with a fixed strategy proposed by \citet{DBLP:conf/iclr/MenonRRK20:PHuberCE}, where $\tau^{(t)}$ remains constant.

The experimental results are presented in Table~\ref{table: adjusting methods}, and the experimental details can be found in Appendix.

As observed, our optimize-based strategy consistently outperforms other approaches in most label noise scenarios, particularly in cases with high noise rates (Sym-80\%) and instance-dependent noise.
Even in scenarios where it does not achieve the highest accuracy, it remains highly competitive, with only a 0.8\% difference compared to the EMA method.
Furthermore, our proposed optimize-based strategy achieves the highest average performance, outperforming the second-best method by 5\%, highlighting its superior overall effectiveness.
These experimental results underscore the robustness and efficacy of our optimize-based strategy across various types of label noise.

\textbf{Can OGC improves robust loss functions?}
Our proposed OGC can be applied not only to non-robust loss functions, such as CE and FL, but also to robust loss functions.
This naturally raises the question of whether an existing robust loss function can improve its performance by incorporating our proposed OGC.
Here, we use Generalized Cross Entropy (GCE) \cite{DBLP:conf/nips/ZhangS18:GCE} as an example and demonstrate that equipping with OGC can enhance the performance of a robust loss function.

Specifically, we conducted a series of experiments on the CIFAR-10 dataset under various types of label noise.
For our GCE+OGC, GCE equipped with OGC, we set the parameters \((q, \epsilon_0)\) to \((0.7, 20.0)\).
The results, which reflect the average accuracies over the last 10 epochs, are presented in Table~\ref{table: GCE+OGC}.
As observed, equipping GCE with our OGC results in consistent performance improvements.
Notably, under asymmetric label noise, we achieved a \(21.01\%\) improvement.
These results demonstrate that our proposed OGC can enhance the effectiveness of existing robust loss functions.

\subsection{Evaluation on Benchmark Datasets}
\label{CIFAR-10 and CIFAR-100}

In this subsection, we evaluate our proposed method using two well-known datasets: CIFAR-10 and CIFAR-100 \cite{krizhevsky2009learning:CIFAR-10/-100}.
To demonstrate the robustness of our proposed method against different types of label noise, we compare it with other state-of-the-art methods under various label noise conditions, including symmetric, asymmetric, instance-dependent \cite{DBLP:conf/nips/XiaL0WGL0TS20:PDN}, and real-world \cite{DBLP:conf/iclr/WeiZ0L0022:CIFAR-10N/-100N} label noise settings.

\textbf{Baselines.}
We consider several state-of-the-art methods:
1) Generalized Cross Entropy (GCE) \cite{DBLP:conf/nips/ZhangS18:GCE};
2) Symmetric Cross Entropy (SCE) \cite{DBLP:conf/iccv/0001MCLY019:SCE};
3) Partially Huberised Cross Entropy (PHuber-CE) \cite{DBLP:conf/iclr/MenonRRK20:PHuberCE};
4) Taylor Cross Entropy (Taylor-CE) \cite{DBLP:conf/ijcai/FengSLL0020:Taylor-CE};
5) Normalized Loss Functions \cite{DBLP:conf/icml/MaH00E020:NormalizedLoss}, specifically NCE+RCE;
6) Jensen-Shannon Divergence Loss (JS) \cite{DBLP:conf/nips/EnglessonA21:GJSLoss};
7) Logit Clipping on Cross Entropy (LC-CE) \cite{DBLP:conf/icml/WeiZXF00L23:LogitClip};
8) Normalized Negative Loss Functions \cite{DBLP:conf/nips/YeLD0ST23:NormalizedNegativeLoss}, specifically NCE+NNCE.
Additionally, we employ Cross Entropy (CE), Focal Loss (FL) \cite{DBLP:conf/iccv/LinGGHD17:FocalLoss}, and Mean Absolute Error (MAE) \cite{DBLP:conf/aaai/GhoshKS17:RobustLoss} for network training.
For our proposed Optimized Gradient Clipping (OGC), we consider two loss functions: 1) CE+OGC and 2) FL+OGC.

\textbf{Experimental details.}
Noise generation, training, and parameter settings are in Appendix.

\textbf{Results.}
The results for CIFAR-10 and CIFAR-100 are presented in Table~\ref{table: cifar10} and Table~\ref{table: cifar100}, respectively. 
The reported values reflect the average accuracies over the last 10 epochs. 
For both datasets, the incorporation of OGC enables CE and FL to consistently achieve the top-2 average accuracy across various scenarios. 
Notably, CE and FL are among the least robust loss functions, emphasizing OGC's effectiveness in enhancing the performance of non-robust loss functions.
Notably, CE and FL, typically among the least robust loss functions, show significant improvement with OGC, highlighting its effectiveness in enhancing the performance of non-robust loss functions. 
For instance, in Table~\ref{table: cifar10}, on CIFAR-10 with real-world label noise, integrating OGC with CE achieves a direct improvement (+17.02\%). 
Moreover, in Table~\ref{table: cifar100}, with the integration of OGC, CE and FL consistently achieve the top two accuracies in all scenarios except for CIFAR-10 with asymmetric label noise. 
Additionally, due to our optimize-based strategy, our loss function requires only a single hyper-parameter, significantly reducing the time needed for hyper-parameter tuning compared to the previous SOTA method, NCE+NNCE, which involves three hyper-parameters.
These observations underscore the effectiveness of our proposed OGC.

\subsection{Evaluation on Real-world Noisy Dataset}

To evaluate our proposed method on large-scale real-world noisy dataset, we conduct experiments on WebVision 1.0 dataset.
WebVision 1.0 \cite{DBLP:journals/corr/abs-1708-02862:WebVision} contains more than 2.4 million web images crawled from the internet by using queries generated from the 1,000 class labels of the ILSVRC 2012 \cite{DBLP:conf/cvpr/DengDSLL009:ImageNet} benchmark.
Following the ``Mini'' setting in previous works \cite{DBLP:conf/icml/JiangZLLF18:MentorNet}, we only take the first 50 classes of the Google resized image subset.
We evaluate the trained networks on the same 50 classes of both the ILSVRC 2012 validation set and WebVision 1.0 validation set, these can be considered as clean validation sets.

The experimental details can be found in Appendix.
The results of last epoch validation accuracies are reported in Table~\ref{table: webvision}.
As observed, integrating our proposed OGC with CE and FL results in significant improvements, outperforming existing robust loss functions on both validation sets.
These results verify that our method is effective in enhancing noise-robustness in large-scale real-world scenarios.

\section{Conclusions}
\label{sec: Conclusions and Limitations}

In this paper, we introduce Optimized Gradient Clipping (OGC), a novel method for noisy label learning, which employs a 2-component Gaussian Mixture Model (2-GMM) to distinguish between clean and noisy samples, estimating the noise-to-clean gradient ratio after clipping.
This allows us to set a clipping threshold at each training step, limiting the influence of noise gradients.
OGC enhances the robustness of both non-robust and robust loss functions in noisy label settings.
Our theoretical analysis highlights its noise tolerance, and extensive experiments demonstrate its effectiveness across various noise types.

\section*{Acknowledgments}

This work is supported by the Shanghai Engineering Research Center of Intelligent Computing System (grant No.~\texttt{19DZ2252600}) and the Research Grants Council (RGC) under grant \texttt{ECS-22303424}.

\bibliography{ref.bib}

\appendix
\onecolumn
\setcounter{secnumdepth}{2}

\setlength{\parindent}{0pt}

\bigskip
\begin{center}
{\LARGE\bf Technical Appendix to ``Optimized Gradient Clipping for Noisy Label Learning''}
\end{center}

\spacingset{1.5}

\section{Method}

\subsection{Algorithm}
\label{appendix: algorithm}

Our algorithm is presented in Algorithm~\ref{alg:AGC}.
For an in-depth discussion on the queue $Q$ and the time frame $s$, please refer to Appendix~\ref{appendix: empirical understanding}.

\begin{algorithm}[ht]
    \small
    \caption{Training process for Optimized Gradient Clipping (OGC)}
    \label{alg:AGC}
    \begin{algorithmic}[1]
    \Require Randomly initialized model \(f\); Dataset \(D_\eta\); FIFO queue \(Q\) with size \(q\); Maximum training steps \(T\); Mini-batch size \(M\); Time frame \(s\); Loss function \(\ell\).
    \For{\(t = 1\) to \(T\)}
        \State Randomly sample a mini-batch \(\{\boldsymbol{x}_m, \tilde{y}_m\}_{m=1}^M\) from \(D_\eta\)
        \For{\(m = 1\) to \(M\)}
            \State \(H_m \gets H(f(\boldsymbol{x}_m), \tilde{y}_m)\) \Comment{Compute cross entropy for sample}
            \State \Call{Enqueue}{$Q, H_m$}
            \Comment{Add cross entropy value to queue}
        \EndFor
        \If{\(t \mod s = 0\)}
            \State \(\mathcal{N}_c, \mathcal{N}_n \gets \text{Fit 2-GMM on } Q\) \Comment{Fit a two-component Gaussian Mixture Model}
            \State \(\tau^{(t)} \gets \text{Solve for optimal }\tau^{(t)} \text{using Equation~(\ref{optimize to get tau_t})}\) \Comment{Update clipping threshold}
        \Else
            \State \(\tau^{(t)} \gets \tau^{(t-1)}\) \Comment{Carry forward the last threshold}
        \EndIf
        \State Compute empirical loss using $\bar{\ell}$
        \Comment{Compute loss with gradient clipping}
        \State Update \(f\) using back-propagation \Comment{Update model parameters}
    \EndFor
    \State \textbf{return} Trained model \(f\)
    \end{algorithmic}
\end{algorithm}

\subsection{Limitations}
While we have illustrated the success of OGC, it is also crucial to understand the limitations that arise in more complex settings.
When the noise distribution is complex or the noise rate is high, the 2-GMM we use may struggle to model the distributions of clean and noise data accurately.
This can lead to unreliable estimates of the ground-truth \( r^{(t)} \), thereby damaging the performance of our proposed OGC.

\section{Experiments}
\label{appendix: experiments}

\subsection{Empirical Understanding}
\label{appendix: empirical understanding}

\textbf{Comparison with manually designed methods.} 
In addition to the discussion in main text, we provide the hyper-parameters used in Table~\ref{table: adjusting methods} for the comparison of manually designed methods.
For the fixed strategy, the results match those of PHuber-CE, as the fixed strategy is equivalent to PHuber-CE.
For the linear strategy, we tuned $\beta \in \{1, 2, 5, 10\}$ and selected $\beta=10$, which performed the best on average.
For the EMA strategy, we tuned $\alpha \in \{0.99, 0.999, 0.9999, 0.99999\}$ and selected $\alpha=0.9999$, which achieved the best performance on average. 

\textbf{Can OGC improves robust loss functions?}
In addition to the discussion in main text, we provide the hyper-parameters used in Table~\ref{table: GCE+OGC}.
For $q$, we adopted the optimal value as reported in the GCE paper \cite{DBLP:conf/nips/ZhangS18:GCE}.
For $\epsilon_0$, we tuned it over $\{100, 50, 20, 10\}$, following the tuning procedure outlined in Appendix~\ref{appendix: evaluation on benchmark datasets}, and selected the value that performed best under 0.5 symmetric label noise.

\begin{figure*}[htbp]
    \centering
    \begin{subfigure}{0.35\textwidth}
        \includegraphics[width=\textwidth]{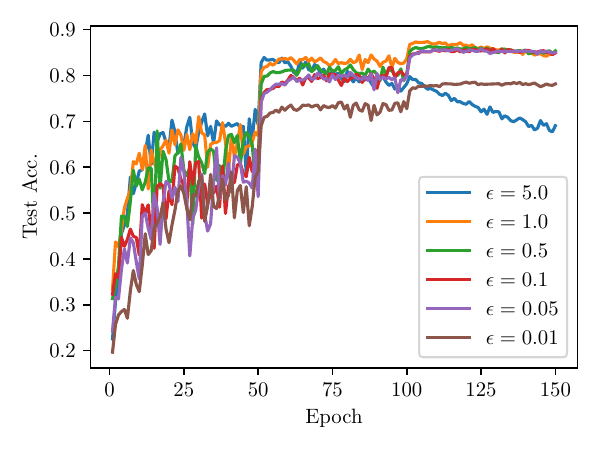}
        \vspace{-1.5em}
        \caption{CIFAR-10, 50\%-Sym.}
    \end{subfigure}
    \hspace{0.05\textwidth}
    \begin{subfigure}{0.35\textwidth}
        \includegraphics[width=\textwidth]{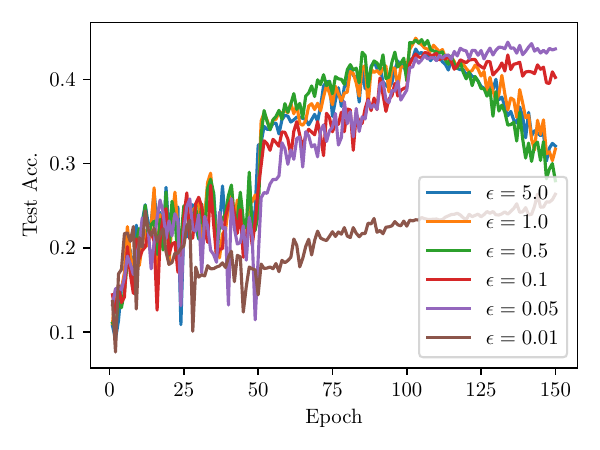}
        \vspace{-1.5em}
        \caption{CIFAR-10, 80\%-Sym.}
    \end{subfigure}
    \caption{Test accuracies (\%) of different $\epsilon$ on CIFAR-10 dataset with different label noise.}
    \label{fig: parameter analysis}
\end{figure*}

\textbf{Predefined gradient ratio $1 + \epsilon$.}
To understand the effect of $1 + \epsilon$, we conduct a series of experiments on the CIFAR-10 dataset under 50\% and 80\% symmetric label noise.
We test various values of $\epsilon \in \{5.0, 1.0, 0.5, 0.1, 0.05, 0.01\}$.
The results, illustrated in Figure~\ref{fig: parameter analysis}, reveal that a smaller $\epsilon$ (e.g., $\epsilon = 0.01$) makes the training robust to noisy labels but leads to underfitting.
Conversely, a larger $\epsilon$ (e.g., $\epsilon = 5.0$) reduces robustness against noisy labels and leads to overfitting.
These findings highlight the necessity of tuning $\epsilon$ to achieve optimal performance.

Regarding parameter tuning, we observe that at each training stage with a different learning rate, the optimal $\epsilon$ varies.
Therefore, we empirically set $\epsilon = \texttt{lr} \cdot \epsilon_0$, where $\epsilon_0$ is the only parameter to tune and $\texttt{lr}$ denotes the learning rate.
By using the same parameter settings across all experiments, we consistently achieve satisfactory results.

\textbf{Queue \(Q\).}
In practice, modeling distributions using a 2-Gaussian Mixture Model (2-GMM) based on a single sampled mini-batch data has the following issues:
1) The resulting Gaussian distribution can exceed the range of cross-entropy values, \(H\);
2) The sample size from a single mini-batch may be insufficient to model reliable distributions.
To address these issues, we employ truncated Gaussian distributions to prevent the values from exceeding the specified range.
Additionally, we maintain a First-In, First-Out (FIFO) queue \(Q\) to accumulate the cross-entropy values from the latest \(q\) samples for modeling the 2-GMM.
The queue size \(q\) influences the number of samples used for modeling the 2-GMM. 

\begin{figure}[t]
    \centering
    \begin{subfigure}{0.35\textwidth}
        \includegraphics[width=\textwidth]{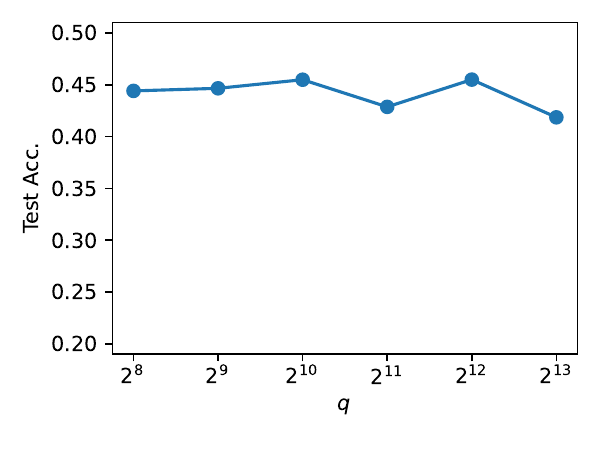}
    \end{subfigure}
    \begin{subfigure}{0.35\textwidth}
        \includegraphics[width=\textwidth]{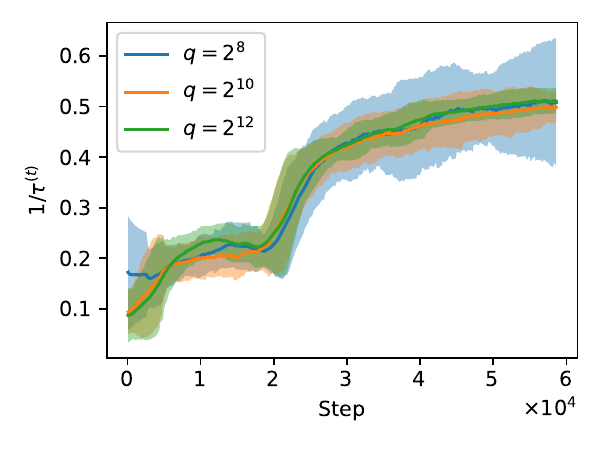}
    \end{subfigure}
    \caption{The effect of queue size \(q\).}
    \label{fig: The effect of queue size q}
\end{figure}

We conduct experiments with \(\epsilon = 0.05\) and \( q \in \{2^8, 2^9, 2^{10}, 2^{11}, 2^{12}, 2^{13}\} \) on CIFAR-10 dataset under 0.8 symmetric label noise.
The results are illustrated in Figure~\ref{fig: The effect of queue size q}.
As observed, while the test accuracies remain nearly consistent across different values of \( q \), the variance of \(\tau^{(t)}\) increases as \( q \) becomes larger.
For all experiments, we set \( q = 2^{12} = 4096 \).

\textbf{Time frame \(s\).}
To solve Eq~\ref{optimize to get tau_t}, we initially apply discretization to divide the range of the cross entropy value \(H\) into predefined intervals, following binary search to locate the optimal \(\tau^{(t)}\).
To enhance efficiency further, we utilize a time frame \(s\) to update \(\tau^{(t)}\), allowing us to avoid solving Eq~\ref{optimize to get tau_t} at each training step \(t\).
The time frame \(s\) controls the frequency of updating \(\tau^{(t)}\).

\begin{figure}[t]
    \centering
    \begin{subfigure}{0.35\textwidth}
        \includegraphics[width=\textwidth]{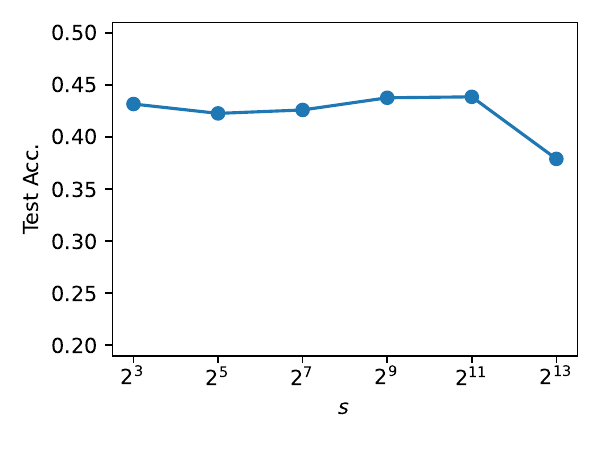}
    \end{subfigure}
    \begin{subfigure}{0.35\textwidth}
        \includegraphics[width=\textwidth]{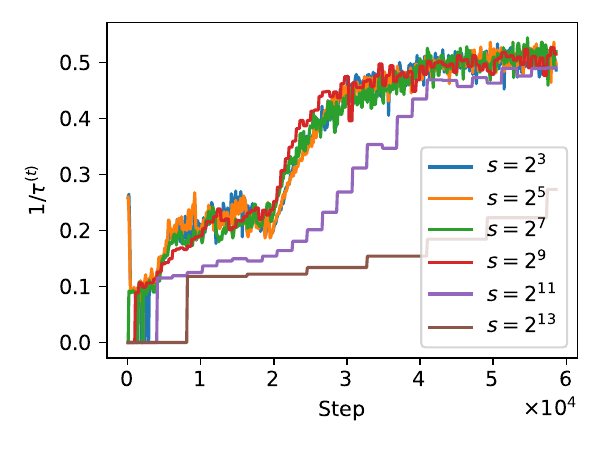}
    \end{subfigure}
    \caption{The effect of interval \(s\).}
    \label{fig: The effect of interval s}
\end{figure}

We conduct experiments with $\epsilon = 0.05$ and \(s \in \{2^3, 2^5, 2^7, 2^9, 2^{11}, 2^{13}\}\) on CIFAR-10 dataset under 0.8 symmetric label noise.
The results are illustrated in Figure~\ref{fig: The effect of interval s}.
As can be observed, when \(s\) becomes larger and exceeds a certain threshold, our \(\tau^{(t)}\) fails to timely capture the model's state, which consequently impairs performance.
In all experiments, we set \(s = 2^5 = 32\) to balance computational efficacy and performance.

\textbf{Time consumption.}
In our method, we employ a 2-GMM to model the distribution of the cross-entropy loss for both clean and noisy samples.
Although a 2-GMM can be considered a non-parametric approach, the additional time required for modeling the distribution could potentially affect the efficiency of our proposed method.
To address this concern, we conduct experiments on a single GTX1080Ti GPU, recording the time consumption and comparing it across different methods.


Specifically, we conduct experiments with CE, Phuber-CE \cite{DBLP:conf/iclr/MenonRRK20:PHuberCE} and CE+OGC on the CIFAR-10 dataset, following the experimental settings outlined in Section~\ref{CIFAR-10 and CIFAR-100}.
The results are summarized in Table~\ref{table: training efficiency}.
As can be observed, compared to CE and Phuber-CE, our proposed method only marginally increases the time cost by 5 minutes (7.14\%) and 3 minutes (4.16\%), respectively.
In contrast, our method shows significantly superior performance.
For instance, it achieves an 18\% improvement over CE under 50\% symmetric label noise and nearly a 13\% improvement over Phuber-CE under 80\% symmetric label noise (see Table~\ref{table: cifar10}).
Overall, the additional time consumption of 2-GMM is minimal, while our method delivers superior performance.

\begin{table}[t]
    \small
    \centering
    \begin{tabular}{cccc}
        \toprule
        Methods & CE & Phuber-CE & CE+OGC \\
        \midrule
        Time consumption (min) & 70 & 72 & 75 \\
        \bottomrule
    \end{tabular}
    \caption{Time consumption for training different methods on the CIFAR-10 dataset.}
    \label{table: training efficiency}
\end{table}

\subsection{Evaluation on Benchmark Datasets}
\label{appendix: evaluation on benchmark datasets}

\textbf{Noise generation.}
For class-independent noise, the noisy labels are generated following standard approaches in previous works \cite{DBLP:conf/icml/MaH00E020:NormalizedLoss, DBLP:conf/icml/ZhouLJGJ21:AsymmetricLoss}.
For symmetric noise, we flip the labels in each class randomly to incorrect labels of other classes.
For asymmetric noise, we flip the labels within a specific set of classes.
For CIFAR-10, flipping TRUCK $\to$ AUTOMOBILE, BIRD $\to$ AIRPLANE, DEER $\to$ HORSE, CAT $\leftrightarrow$ DOG.
For CIFAR-100, the 100 classes are grouped into 20 super-classes with each has 5 sub-classes, and each class are flipped within the same super-class into the next in a circular fashion.
We vary the noise rate $\eta \in \{0.2,0.5,0.8\}$ for symmetric noise and $\eta \in \{0.4\}$ for asymmetric noise.
For instance-dependent noise, we use the part-dependent noise from PDN \cite{DBLP:conf/nips/XiaL0WGL0TS20:PDN} with a noise rate of 40\%, where the noise is synthesized based on the DNN prediction error.
For real-world noise, we use the “Worst" label set of CIFAR-10N and the “Noisy" label set of CIFAR-100N \cite{DBLP:conf/iclr/WeiZ0L0022:CIFAR-10N/-100N}, respectively.

\textbf{Training details.}
We employ a ResNet-18 for CIFAR-10 and a ResNet-34 \cite{DBLP:conf/cvpr/HeZRS16:ResNet} for CIFAR-100.
For both CIFAR-10 and CIFAR-100, the networks are trained for 150 epochs.
For all the training, we use SGD optimizer with 0.9 momentum.
We set the initial learning rate as 0.1, and reduce it by a factor of 10 after 50 and 100 epochs.
Weight decay is set to $5 \times 10^{-4}$.
Batch size is set to $128$.
For all settings, we clip the gradient norm to $5.0$.
Typical data augmentations including random horizontal flip, random resized crop and random rotation are applied.
We conduct all the experiments on NVIDIA GeForce RTX 3090, and implement all methods by PyTorch.

\textbf{Parameter settings.}
For all baseline methods, we tuned their parameters according to the guidelines provided in their original papers.
For both CIFAR-10 and CIFAR-100 datasets, tuning was performed using 50\% symmetric label noise to obtain the optimal parameters, which were then applied consistently across all other label noise settings.
To ensure a fair comparison, we employed an identical grid search approach to find the optimal parameters for each method within a comparable budget.
Specifically, we tuned $q \in \{0.9, 0.8, 0.7, 0.6, 0.5, 0.4\}$ for GCE, $\alpha \in \{0.1, 0.5, 1.0, 5.0, 10.0\}$ and $\beta \in \{0.1, 0.5, 1.0, 5.0, 10.0\}$ for SCE, $\tau \in \{2, 3, 4, 5, 6, 7, 8, 9, 10\}$ for PHuber-CE, $t \in \{2, 4, 6, 8, 10, 12, 14, 16, 18\}$ for Taylor-CE, $\alpha \in \{0.1, 1.0, 10.0, 50.0, 100.0\}$ and $\beta \in \{0.01, 0.1, 1.0, 10.0\}$ for NCE+RCE, $\pi_1 \in \{0.1, 0.2, 0.3, 0.4, 0.5, 0.6, 0.7, 0.8, 0.9\}$ for JS, $\tau \in \{10, 2, 1, 0.5, 0.2, 0.1, 0.05, 0.02, 0.01\}$ for LC-CE, and $\alpha \in \{0.1, 1.0, 10.0, 50.0, 100.0\}$ and $\beta \in \{0.1, 1.0, 5.0, 10.0\}$ for NCE+NNCE.
For our proposed OGC, we tuned $\epsilon_0 \in \{100.0, 50.0, 20.0, 10.0, 5.0, 1.0\}$.
Detailed parameter configurations are presented in Table~\ref{table: param}.

\subsection{Evaluation on Real-world Noisy Dataset}
\label{appendix: evaluation on real-world noisy dataset}

\textbf{Training details.}
We follow the experimental settings in previous work \cite{DBLP:conf/icml/WeiZXF00L23:LogitClip} and train a ResNet-18 network using SGD for 120 epochs with an initial learning rate of 0.1, Nesterov momentum 0.9, weight decay \(5 \times 10^{-4}\), and batch size 128.
The learning rate is reduced by a factor of 10 after 40 and 80 epochs.
We resize the images to \(224 \times 224\) and apply the standard data augmentations, including random cropping and random horizontal flip.

\textbf{Parameter settings.}
For all baseline methods, we mainly followed the parameter settings in \citet{DBLP:conf/icml/WeiZXF00L23:LogitClip}.
For our proposed OGC, we tuned $\epsilon_0 \in \{400.0, 350.0, 300.0, 250.0, 200.0, 150.0, 100.0\}$ to find the optimal parameters.
Detailed parameter configurations are presented in Table~\ref{table: param}.

\begin{table}[h]
\centering
\small
\begin{tabular}{cccc}
    \toprule
    Methods & CIFAR-10 & CIFAR-100 & WebVision \\
    \midrule
    CE (-) & (-) & (-) & (-) \\
    FL ($\gamma$) & (\(0.5\)) & (\(0.5\)) & - \\
    MAE (-) & (-) & (-) & - \\
    GCE (\(q\)) & (\(0.9\)) & (\(0.6\)) & (\(0.7\)) \\
    SCE (\(\alpha, \beta\)) & (\(0.1, 1.0\)) & (\(0.1, 1.0\)) & ($0.5, 1.0$) \\
    PHuber-CE (\(\tau\)) & (\(2\)) & (\(10\)) & (\(30\)) \\
    Taylor-CE (\(t\)) & (\(2\)) & (\(16\)) & - \\
    NCE+RCE (\(\alpha, \beta\)) & (\(1.0, 0.1\)) & (\(50.0, 0.1\)) & (\(50.0, 0.1\)) \\
    JS (\(\pi_1\)) & (\(0.9\)) & (\(0.5\)) & - \\
    LC-CE (\(\tau\)) & ($0.1$) & ($0.1$) & ($1.2$) \\
    NCE+NNCE (\(\alpha, \beta\)) & (\(1.0, 1.0\)) & (\(100.0, 5.0\)) & - \\
    \midrule
    CE+OGC (\(\epsilon_0\)) & (\(20.0\)) & (\(20.0\)) & (\(250.0\)) \\
    FL+OGC (\(\gamma, \epsilon_0\)) & (\(0.5, 20.0\)) & (\(0.5, 20.0\)) &(\(0.5, 350.0\)) \\
    \bottomrule
\end{tabular}
\caption{Parameter settings for different methods.}
\label{table: param}
\end{table}

\section{Proofs}
\label{appendix: proofs}

Our proofs are inspired by \citet{DBLP:conf/nips/ZhangS18:GCE} and \citet{DBLP:conf/icml/WeiZXF00L23:LogitClip}.

\subsection{Proof of Proposition~\ref{proposition: bounded loss}}
\label{proof: bounded loss}

\begin{proof}
    We first consider $p(j|x) \in [0, \frac{1}{\tau^{(t)}}]$.
    For upper bound, we have:
    \begin{align}
        & \quad \ (1 - p(j|x))(1 + \log \tau^{(t)}) - \bar{\ell}_\text{CE}(f(\bm x), j, \tau^{(t)}) \\
        & = (1 - p(j|x))(1 + \log \tau^{(t)}) - (1 - \tau^{(t)} \cdot p(j|x) + \log \tau^{(t)}) \\
        & = p(j|x) (\tau^{(t)} - 1 - \log \tau^{(t)}) \\
        & \ge 0,
    \end{align}
    the last inequality holds because $p(j|x) \ge 0$, $\tau^{(t)} - 1 - \log \tau^{(t)}$ is increasing for $\tau^{(t)} \ge 1$ and $\tau^{(t)} - 1 - \log \tau^{(t)} = 0$ when $\tau^{(t)} = 1$. 
    For lower bound, we have:
    \begin{align}
        & \quad \ \bar{\ell}_\text{CE}(f(\bm x), j, \tau^{(t)}) - (1 - p(j|x)) \\
        & = (1 - \tau^{(t)} \cdot p(j|x) + \log \tau^{(t)}) - (1 - p(j|x)) \\
        & = p(j|x) (1 - \tau^{(t)}) + \log \tau^{(t)} \\
        & \ge \frac{1}{\tau^{(t)}} (1 - \tau^{(t)}) + \log \tau^{(t)} \\
        & = \frac{1}{\tau^{(t)}} - 1 + \log \tau^{(t)} \\
        & \ge 0,
    \end{align}
    the last inequality holds because $\frac{1}{\tau^{(t)}} - 1 + \log \tau^{(t)}$ is increasing for $\tau^{(t)} \ge 1$, and $\frac{1}{\tau^{(t)}} - 1 + \log \tau^{(t)} = 0$ when $\tau^{(t)} = 1$.

    Next, we consider $p(j|x) \in (\frac{1}{\tau^{(t)}}, 1]$.
    For upper bound, we have:
    \begin{align}
        & \quad \ (1 - p(j|x))(1 + \log \tau^{(t)}) - \bar{\ell}_\text{CE}(f(\bm x), j, \tau^{(t)}) \\
        & = (1 - p(j|x))(1 + \log \tau^{(t)}) - (- \log p(j|x)) \\
        & \ge 0,
    \end{align}
    the last inequality holds because $(1 - p(j|x))(1 + \log \tau^{(t)}) - (- \log p(j|x))$ is decreasing for $p(j|x) \le 1$ and $\tau^{(t)} \ge 1$, and $(1 - p(j|x))(1 + \log \tau^{(t)}) - (- \log p(j|x)) = 0$ when $p(j|x) = 1$.
    For lower bound, we have:
    \begin{align}
        & \quad \ \bar{\ell}_\text{CE}(f(\bm x), j, \tau^{(t)}) - (1 - p(j|x)) \\
        & = - \log p(j|x) - (1 - p(j|x)) \\
        & \ge 0,
    \end{align}
    the last inequality holds because $- \log p(j|x) - (1 - p(j|x))$ is decreasing for $p(j|x) \le 1$, and $- \log p(j|x) - 1 + p(j|x) = 0$ when $p(j|x) = 1$.
    
    Finally, for $p(j|x) \in [0, 1]$ and $\tau^{(t)} \ge 1$, we have:
    \begin{equation}
        1 - p(j|x) \le \bar{\ell}_\text{CE}(f(\bm x), j, \tau^{(t)}) \le (1 - p(j|x))(1 + \log \tau^{(t)}),
    \end{equation}
    and,
    \begin{equation}
        K - 1 \le \sum_{j=1}^K \bar{\ell}_\text{CE}(f(\bm x), j, \tau^{(t)}) \le (K - 1)(1 + \log \tau^{(t)}).
    \end{equation}
    Moreover, when $\tau^{(t)} = 1$, the upper bound $(K - 1)(1 + \log \tau^{(t)})$ is equal to the lower bound $K - 1$, thus,
    \begin{equation}
        \sum_{j=1}^K \bar{\ell}_\text{CE}(f(\bm x), j, \tau^{(t)}) = K - 1, \quad \text{when} \quad \tau^{(t)} = 1.
    \end{equation}
    This completes the proof.
\end{proof}

\subsection{Proof of Theorem~\ref{theorem: noise-tolerant under symmetric noise}}
\label{proof: noise-tolerant under symmetric noise}

\begin{proof}
    Recall that for symmetric noise, we have $q_{\eta}(j|y) = \eta_{yj} = \frac{\eta}{K - 1}$, where $\eta$ is the noise rate. Then, given any model $f$ and $\tau^{(t)} \ge 1$,
    \begin{align}
        \mathcal{R}^\eta_{\bar{\ell}_\text{CE}}(f, \tau^{(t)})
        & = \mathbb{E}_{(\bm x, \tilde{y}) \sim \mathcal{D}_\eta} \big[ \bar{\ell}_\text{CE}(f(\bm x), \tilde{y}, \tau^{(t)}) \big] \\
        & = \mathbb{E}_{(\bm x, y) \sim \mathcal{D}_c} \mathbb{E}_{\tilde{y}|\bm x, y} \big[ \bar{\ell}_\text{CE}(f(\bm x), \tilde{y}, \tau^{(t)}) \big] \\
        & = \mathbb{E}_{(\bm x, y) \sim \mathcal{D}_c} \Big[ (1 - \eta) \bar{\ell}_\text{CE}(f(\bm x), y, \tau^{(t)}) + \sum_{j \ne y} \frac{\eta}{K - 1} \bar{\ell}_\text{CE}(f(\bm x), j, \tau^{(t)}) \Big] \\
        & = \mathbb{E}_{(\bm x, y) \sim \mathcal{D}_c} \Big[ (1 - \frac{\eta K}{K - 1}) \bar{\ell}_\text{CE}(f(\bm x), y, \tau^{(t)}) + \frac{\eta}{K - 1} \sum_j^K \bar{\ell}_\text{CE}(f(\bm x), j, \tau^{(t)}) \Big] \\
        & = \Big( 1 - \frac{\eta K}{K - 1} \Big) \mathcal{R}_{\bar{\ell}_\text{CE}}(f, \tau^{(t)}) +  \frac{\eta}{K - 1} \mathbb{E}_{(\bm x, y) \sim \mathcal{D}_c} \Big[ \sum_{j=1}^K \bar{\ell}_\text{CE}(f(\bm x), j, \tau^{(t)}) \Big].
    \end{align}
    From Proposition~\ref{proposition: bounded loss}, we have:
    \begin{equation}
        \Big( 1 - \frac{\eta K}{K - 1} \Big) \mathcal{R}_{\bar{\ell}_\text{CE}}(f, \tau^{(t)}) + \eta \le \mathcal{R}^\eta_{\bar{\ell}_\text{CE}}(f, \tau^{(t)}) \le \Big( 1 - \frac{\eta K}{K - 1} \Big) \mathcal{R}_{\bar{\ell}_\text{CE}}(f, \tau^{(t)}) + \eta (1 + \log \tau^{(t)}).
    \end{equation}
    We can also write the inequality in terms of $\mathcal{R}_{\bar{\ell}_\text{CE}}(f, \tau^{(t)})$:
    {
    \small
    \begin{equation}
        \Big( \mathcal{R}^\eta_{\bar{\ell}_\text{CE}}(f, \tau^{(t)}) - \eta (1 + \log \tau^{(t)}) \Big) \Big/ \Big( 1 - \frac{\eta K}{K - 1} \Big)
        \le \mathcal{R}_{\bar{\ell}_\text{CE}}(f, \tau^{(t)})
        \le \Big( \mathcal{R}^\eta_{\bar{\ell}_\text{CE}}(f, \tau^{(t)}) - \eta \Big) \Big/ \Big( 1 - \frac{\eta K}{K - 1} \Big)
    \end{equation}
    }
    
    Thus, for $\tilde{f}^\star$,
    \begin{align}
        & \quad \ \mathcal{R}_{\bar{\ell}_\text{CE}}(\tilde{f}^\star, \tau^{(t)}) - \mathcal{R}_{\bar{\ell}_\text{CE}}(f^\star, \tau^{(t)}) \\
        & \le \Big( \mathcal{R}^\eta_{\bar{\ell}_\text{CE}}(\tilde{f}^\star, \tau^{(t)}) - \eta - \mathcal{R}^\eta_{\bar{\ell}_\text{CE}}(f^\star, \tau^{(t)}) + \eta (1 + \log \tau^{(t)}) \Big) \Big/ \Big( 1 - \frac{\eta K}{K - 1} \Big) \\
        & = \Big( \mathcal{R}^\eta_{\bar{\ell}_\text{CE}}(\tilde{f}^\star, \tau^{(t)}) - \mathcal{R}^\eta_{\bar{\ell}_\text{CE}}(f^\star, \tau^{(t)}) + \log \tau^{(t)} \Big) \Big/ \Big( 1 - \frac{\eta K}{K - 1} \Big) \\
        & \le \log \tau^{(t)} \Big/ \Big( 1 - \frac{\eta K}{K - 1} \Big),
    \end{align}
    where $1 - \frac{\eta K}{K - 1} > 0$, since $\eta < \frac{K - 1}{K}$.
    The last inequality holds because when $\tilde{f}^\star$ is a minimizer of $\mathcal{R}^\eta_{\bar{\ell}_\text{CE}}(f, \tau^{(t)})$, we have that $\mathcal{R}^\eta_{\bar{\ell}_\text{CE}}(\tilde{f}^\star, \tau^{(t)}) - \mathcal{R}^\eta_{\bar{\ell}_\text{CE}}(f^\star, \tau^{(t)}) \le 0$.
    Similar, when $f^\star$ is a minimizer of $\mathcal{R}_{\bar{\ell}_\text{CE}}(f, \tau^{(t)})$, we have $0 \le \mathcal{R}_{\bar{\ell}_\text{CE}}(\tilde{f}^\star, \tau^{(t)}) - \mathcal{R}_{\bar{\ell}_\text{CE}}(f^\star, \tau^{(t)})$.
    
    In summary,
    \begin{equation}
        0
        \le \mathcal{R}_{\bar{\ell}_\text{CE}}(\tilde{f}^\star, \tau^{(t)}) - \mathcal{R}_{\bar{\ell}_\text{CE}}(f^\star, \tau^{(t)})
        \le \log \tau^{(t)} \Big/ \Big( 1 - \frac{\eta K}{K - 1} \Big).
    \end{equation}
    This completes the proof.
\end{proof}

\subsection{Proof of Theorem~\ref{theorem: noise-tolerant under asymmetric noise}}
\label{proof: noise-tolerant under asymmetric noise}

\begin{proof}
    Recall that for asymmetric label noise, we have $\sum_{j=1}^K \eta_{yj} = 1$ and $\eta_y = \sum_{j \ne y} \eta_{yj}$,
    \begin{align}
        \mathcal{R}^\eta_{\bar{\ell}_\text{CE}}(f, \tau^{(t)})
        & = \mathbb{E}_{(\bm x, \tilde{y}) \sim \mathcal{D}_\eta} \big[ \bar{\ell}_\text{CE}(f(\bm x), \tilde{y}, \tau^{(t)}) \big] \\
        & = \mathbb{E}_{(\bm x, y) \sim \mathcal{D}_c} \mathbb{E}_{\tilde{y}|\bm x, y} \big[ \bar{\ell}_\text{CE}(f(\bm x), \tilde{y}, \tau^{(t)}) \big] \\
        & = \mathbb{E}_{(\bm x, y) \sim \mathcal{D}_c}[(1 - \eta_y) \bar{\ell}_\text{CE}(f(\bm x), y, \tau^{(t)})] + \mathbb{E}_{(\bm x, y) \sim \mathcal{D}_c} \Big[ \sum_{j \ne y} \eta_{yj} \bar{\ell}_\text{CE}(f(\bm x), j, \tau^{(t)}) \Big].
    \end{align}
    From Proposition~\ref{proposition: bounded loss}, we have:
    \begin{align}
        \mathcal{R}^\eta_{\bar{\ell}_\text{CE}}(f, \tau^{(t)})
        & \le \mathbb{E}_{(\bm x, y) \sim \mathcal{D}_c}\Big[(1 - \eta_y) \Big( (K - 1)(1 + \log \tau^{(t)}) - \sum_{j \ne y} \bar{\ell}_\text{CE}(f(\bm x), j, \tau^{(t)}) \Big) \Big] \\
        & \quad \ + \mathbb{E}_{(\bm x, y) \sim \mathcal{D}_c} \Big[ \sum_{j \ne y} \eta_{yj} \bar{\ell}_\text{CE}(f(\bm x), j, \tau^{(t)}) \Big] \\
        & = (K - 1)(1 + \log \tau^{(t)}) \mathbb{E}_{(\bm x, y) \sim D_c} [1 - \eta_y] \\
        & \quad \ - \mathbb{E}_{(\bm x, y) \sim \mathcal{D}_c} \Big[ \sum_{j \ne y} (1 - \eta_y - \eta_{yj}) \bar{\ell}_\text{CE}(f(\bm x), j, \tau^{(t)}) \Big],
    \end{align}
    and,
    \begin{align}
        \mathcal{R}^\eta_{\bar{\ell}_\text{CE}}(f, \tau^{(t)})
        & \ge \mathbb{E}_{(\bm x, y) \sim \mathcal{D}_c}\Big[(1 - \eta_y) \Big( (K - 1) - \sum_{j \ne y} \bar{\ell}_\text{CE}(f(\bm x), j, \tau^{(t)}) \Big) \Big] \\
        & \quad \ + \mathbb{E}_{(\bm x, y) \sim \mathcal{D}_c} \Big[ \sum_{j \ne y} \eta_{yj} \bar{\ell}_\text{CE}(f(\bm x), j, \tau^{(t)}) \Big] \\
        & = (K - 1) \mathbb{E}_{(\bm x, y) \sim D_c} [1 - \eta_y] \\
        & \quad \ - \mathbb{E}_{(\bm x, y) \sim \mathcal{D}_c} \Big[ \sum_{j \ne y} (1 - \eta_y - \eta_{yj}) \bar{\ell}_\text{CE}(f(\bm x), j, \tau^{(t)}) \Big].
    \end{align}
    Hence, for $\tilde{f}^\star$,
    \begin{align}
        & \quad \ \mathcal{R}^\eta_{\bar{\ell}_\text{CE}}(f^\star, \tau^{(t)}) - \mathcal{R}^\eta_{\bar{\ell}_\text{CE}}(\tilde{f}^\star, \tau^{(t)}) \\
        & \le (K - 1)(\log \tau^{(t)}) \mathbb{E}_{(\bm x, y) \sim D_c} [1 - \eta_y] \\
        & \quad \ + \mathbb{E}_{(\bm x, y) \sim \mathcal{D}_c} \Big[ \sum_{j \ne y} (1 - \eta_y - \eta_{yj}) \Big( \bar{\ell}_\text{CE}(\tilde{f}^\star(\bm x), j, \tau^{(t)}) - \bar{\ell}_\text{CE}(f^\star(\bm x), j, \tau^{(t)}) \Big) \Big].
    \end{align}
    Now, from our assumption that $\mathcal{R}_{\bar{\ell}_\text{CE}}(f^\star, \tau^{(t)}) = 0$, we have $\bar{\ell}_\text{CE}(f^\star(\bm x), y, \tau^{(t)}) = 0$.
    This is only satisfied iff $p(y|x)=1$ and $p(j|x) = 0 , \forall j \ne y$.
    As a result, $\bar{\ell}_\text{CE}(f^\star(\bm x), j, \tau^{(t)}) = 1 + \log \tau^{(t)}, \forall j \ne y$.
    Note that from Proposition~\ref{proposition: bounded loss}, we have 
    $$\bar{\ell}_\text{CE}(\tilde{f}^\star(\bm x), j, \tau^{(t)}) \le (1 - p(j|x))(1 + \log \tau^{(t)}) \le 1 + \log \tau^{(t)}, \forall j \ne y.$$
    Moreover, per the assumption $1 - \eta_y - \eta_{yj} \ge 0$, we have
    \begin{equation}
        \sum_{j \ne y} (1 - \eta_y - \eta_{yj}) \Big( \bar{\ell}_\text{CE}(\tilde{f}^\star(\bm x), j, \tau^{(t)}) - \bar{\ell}_\text{CE}(f^\star(\bm x), j, \tau^{(t)}) \Big) \le 0.
    \end{equation}
    Recall that when $\tilde{f}^\star$ is a minimizer of $\mathcal{R}^\eta_{\bar{\ell}_\text{CE}}(f, \tau^{(t)})$, we have $0 \le \mathcal{R}^\eta_{\bar{\ell}_\text{CE}}(f^\star, \tau^{(t)}) - \mathcal{R}^\eta_{\bar{\ell}_\text{CE}}(\tilde{f}^\star, \tau^{(t)})$.
    Finally,
    \begin{equation}
        0 \le \mathcal{R}^\eta_{\bar{\ell}_\text{CE}}(f^\star, \tau^{(t)}) - \mathcal{R}^\eta_{\bar{\ell}_\text{CE}}(\tilde{f}^\star, \tau^{(t)}) \le (K - 1)(\log \tau^{(t)}) \mathbb{E}_{(\bm x, y) \sim D_c} [1 - \eta_y].
    \end{equation}
    This completes the proof.
\end{proof}

\subsection{Proof of Theorem~\ref{theorem: noise-tolerant under instance-dependent noise}}
\label{proof: noise-tolerant under instance-dependent noise}

\begin{proof}
    For instance-dependent noise, we have
    \begin{align}
        \mathcal{R}^\eta_{\bar{\ell}_\text{CE}}(f, \tau^{(t)})
        & = \mathbb{E}_{(\bm x, \tilde{y}) \sim \mathcal{D}_\eta} \big[ \bar{\ell}_\text{CE}(f(\bm x), \tilde{y}, \tau^{(t)}) \big] \\
        & = \mathbb{E}_{(\bm x, y) \sim \mathcal{D}_c} \mathbb{E}_{\tilde{y}|\bm x, y} \big[ \bar{\ell}_\text{CE}(f(\bm x), \tilde{y}, \tau^{(t)}) \big] \\
        & = \mathbb{E}_{(\bm x, y) \sim \mathcal{D}_c}[(1 - \eta_y(\bm x)) \bar{\ell}_\text{CE}(f(\bm x), y, \tau^{(t)})] \\
        & \quad \ + \mathbb{E}_{(\bm x, y) \sim \mathcal{D}_c} \Big[ \sum_{j \ne y} \eta_{yj}(\bm x) \bar{\ell}_\text{CE}(f(\bm x), j, \tau^{(t)}) \Big].
    \end{align}
    From Proposition~\ref{proposition: bounded loss}, we have:
    \begin{align}
        \mathcal{R}^\eta_{\bar{\ell}_\text{CE}}(f, \tau^{(t)})
        & \le (K - 1)(1 + \log \tau^{(t)}) \mathbb{E}_{(\bm x, y) \sim D_c} [1 - \eta_y(\bm x)] \\
        & \quad \ - \mathbb{E}_{(\bm x, y) \sim \mathcal{D}_c} \Big[ \sum_{j \ne y} (1 - \eta_y(\bm x) - \eta_{yj}(\bm x)) \bar{\ell}_\text{CE}(f(\bm x), j, \tau^{(t)}) \Big],
    \end{align}
    and,
    \begin{align}
        \mathcal{R}^\eta_{\bar{\ell}_\text{CE}}(f, \tau^{(t)})
        & \ge (K - 1) \mathbb{E}_{(\bm x, y) \sim D_c} [1 - \eta_y(\bm x)] \\
        & \quad \ - \mathbb{E}_{(\bm x, y) \sim \mathcal{D}_c} \Big[ \sum_{j \ne y} (1 - \eta_y(\bm x) - \eta_{yj}(\bm x)) \bar{\ell}_\text{CE}(f(\bm x), j, \tau^{(t)}) \Big].
    \end{align}
    Hence, for $\tilde{f}^\star$,
    \begin{align}
        & \quad \ \mathcal{R}^\eta_{\bar{\ell}_\text{CE}}(f^\star, \tau^{(t)}) - \mathcal{R}^\eta_{\bar{\ell}_\text{CE}}(\tilde{f}^\star, \tau^{(t)}) \\
        & \le (K - 1)(\log \tau^{(t)}) \mathbb{E}_{(\bm x, y) \sim D} [1 - \eta_y(\bm x)] \\
        & \quad \ + \mathbb{E}_{(\bm x, y) \sim \mathcal{D}_c} \Big[ \sum_{j \ne y} (1 - \eta_y(\bm x) - \eta_{yj}(\bm x)) \Big( \bar{\ell}_\text{CE}(\tilde{f}^\star(\bm x), j, \tau^{(t)}) - \bar{\ell}_\text{CE}(f^\star(\bm x), j, \tau^{(t)}) \Big) \Big].
    \end{align}
    From the Proof~\ref{proof: noise-tolerant under asymmetric noise} (proof of Theorem~\ref{theorem: noise-tolerant under asymmetric noise}), when $\mathcal{R}_{\bar{\ell}_\text{CE}}(f^\star, \tau^{(t)}) = 0$, we have $\bar{\ell}_\text{CE}(\tilde{f}^\star(\bm x), j, \tau^{(t)}) - \bar{\ell}_\text{CE}(f^\star(\bm x), j, \tau^{(t)}) \le 0$.
    Recall that $\eta_{yj}(\bm x) < 1 - \eta_y(\bm x)$, we have
    \begin{equation}
        \sum_{j \ne y} (1 - \eta_y(\bm x) - \eta_{yj}(\bm x)) \Big( \bar{\ell}_\text{CE}(\tilde{f}^\star(\bm x), j, \tau^{(t)}) - \bar{\ell}_\text{CE}(f^\star(\bm x), j, \tau^{(t)}) \Big) \le 0.
    \end{equation}
    Finally,
    \begin{equation}
        0
        \le \mathcal{R}^\eta_{\bar{\ell}_\text{CE}}(f^\star, \tau^{(t)}) - \mathcal{R}^\eta_{\bar{\ell}_\text{CE}}(\tilde{f}^\star, \tau^{(t)})
        \le (K - 1)(\log \tau^{(t)}) \mathbb{E}_{(\bm x, y) \sim D_c} [1 - \eta_y(\bm x)].
    \end{equation}
    This completes the proof.
\end{proof}

\end{document}